\documentclass[a4paper,conference]{IEEEtran}
\usepackage{cite}

\ifCLASSINFOpdf
   \usepackage[pdftex]{graphicx}
\else
\fi
\usepackage{amsmath}
\usepackage{algpseudocode} 

\usepackage{userDefinedCommands}

\begin{document}
%
\title{K-Means for Noise-Insensitive\\ Multi-Dimensional Feature Learning}

\author{
\IEEEauthorblockN{Nicholas Pellegrino \hspace*{0.6in} Paul W. Fieguth \hspace*{0.6in} Parsin Haji Reza}
\IEEEauthorblockA{PhotoMedicine Labs, Department of Systems Design Engineering, University of Waterloo\\
Waterloo, Ontario, Canada\\
npellegr@uwaterloo.ca \qquad pfieguth@uwaterloo.ca \qquad phajireza@uwaterloo.ca}
}


%



\maketitle

\begin{abstract}
Many measurement modalities which perform imaging by probing an object pixel-by-pixel, such as via Photoacoustic Microscopy, produce a multi-dimensional feature (typically a time-domain signal) at each pixel.  In principle, the many degrees of freedom in the time-domain signal would admit the possibility of significant multi-modal information being implicitly present, much more than a single scalar ``brightness'', regarding the underlying targets being observed. However, the measured signal is neither a weighted-sum of basis functions (such as principal components) nor one of a set of prototypes (\mbox{K-means}), which has motivated the novel clustering method proposed here.
Signals are clustered based on their shape, but not amplitude, via angular distance 
and centroids are calculated as the direction of maximal intra-cluster variance, resulting in a clustering algorithm capable of learning centroids (signal shapes) that are related to the underlying, albeit unknown, target characteristics in a scalable and noise-robust manner.
\end{abstract}


%
\IEEEpeerreviewmaketitle

\section{Introduction}
\label{sec:intro}

Medical imaging broadly serves as a powerful diagnostic tool, often non-invasively giving medical care providers valuable information that cannot be obtained otherwise \cite{burbridge2017undergraduate,wang2012biomedical}.
Many biomedical sensing techniques operate by capturing time-domain (TD) signals from which diagnostically relevant information can be inferred. A list of some of these modalities includes sonography / ultrasound imaging \cite{allan2011clinical,hobbins2008obstetric}, echocardiography \cite{laurenceau2012essentials}, electrocardiography \cite{foster2007twelve,wang2012atlas,chugh2014textbook}, electromyography \cite{leis2013atlas,kamen2009essentials}, phonocardiography \cite{tavel1972clinical,leatham1970auscultation}, phonomyography \cite{fuchs2011neuromuscular}, etc. 
In many cases objects are scanned, pixel-by-pixel, to produce a TD signal at each pixel.
Two such modalities are Time-Domain Optical Coherence Tomography (TD-OCT) \cite{haberland1997optical,watanabe2009high} and Photoacoustic Microscopy (PAM) \cite{beard2011biomedical,li2013reflection,yao2013photoacoustic}.

In principal, the many degrees of freedom available within TD signals admit the possibility of significant multi-modal information related to the imaged target being implicitly present, far beyond a single scalar value used to represent pixel ``brightness''. 
However, extracting the information from these signals is not necessarily straightforward. Indeed, the work presented in this paper is motivated by the TD signals of Photoacoustic Remote Sensing (PARS) Microscopy \cite{hajireza2017non,abbasi2019all,abbasi2020rapid,ecclestone2020improving,ecclestone2021histopathology}, a novel all-optical variation of photoacoustic microscopy. The underlying physics lead to TD signals having shapes specific to tissue type, but where the signals from a given target may vary in amplitude, be inverted (negative amplitudes), and suffer from noise. 
What is required is a set of time-domain \emph{features}, that adequately capture information from the underlying target that is present in the TD signals.

\section{Background}
\label{sec:background}

This work is motivated by imaging modalities that scan, pixel-by-pixel, measuring a signal, $\TDInd{j}$, over time, $t$, at each pixel, $j$:
\begin{equation}
    \TDInd{j} = \alpha_j f_i(t),
    \label{eq:model}
\end{equation}
for some weight, $\alpha_j$, applied to feature $f_i$ for target type $i$.  In principle $t$ is continuous, however in practice the TD signals are sampled by a data acquisition system. For the purpose of this work, the same notation is used for both the \mbox{discrete- and} continuous-time representations, however all numerical computations involving the TD signals clearly refer to the measured discrete-time representation.

Conventionally, in PAM and PARS \cite{wang2012biomedical,hajireza2017non,abbasi2019all,abbasi2020rapid,bost2009developing,ku2010photoacoustic,rao2010hybrid}, only a scalar amplitude is extracted from each TD signal, accomplished by either using a Hilbert transform \cite{cizek1970discrete} to find an envelope of the signal, from which the difference between maximum and minimum values is computed, or by directly computing the difference between maximum and minimum of the raw TD signal itself. Recently, other methods have been developed to extract additional information related to the frequency content of the signals, as a means of inferring information related to the imaged target \cite{moore2019photoacousticF_mode,kedarisetti2021label,kedarisetti2021f}. This work takes a different approach by proposing an unsupervised (clustering) approach to learn time-domain features that relate to the underlying  target.

In principle, such a feature inference would seem to have been solved.  Principal Components Analysis (PCA) and its variations \cite{pearson1901liii,hotelling1933analysis,bro2014principal,shlens2014tutorial,hasan2021review}  are capable of extracting features (the principal components) from TD data; the principal components yield a representation,
\begin{equation}
    \TDInd{j} = \sum_i \alpha_{j,i} b_i(t),
\end{equation}
based on a weighted sum of basis elements $b_i$.  However the optimal basis elements are those minimizing the variance of the residual error, and not necessarily those which \emph{individually} effectively represent most of the signals, as in \Cref{eq:model}.  That is, the principal components are unknown weighted combinations of the desired features, and therefore do not individually lead to meaningful features.

Clustering methods, such as K-Means \cite{macqueen1967some,steinhaus1956division,hartigan1979algorithm,lloyd1982least} and K-Medoids \cite{kaufman1990partitioning,schubert2021fast}, assume \emph{all} measured data-points (TD signals) are \emph{one} of a set of prototypes, and have variation only as a consequence of measurement error and noise. This assumption does not hold in \Cref{eq:model}, as measured TD signals may be scaled or inverted versions of what would be a prototype relating to a specific target.
Furthermore, as will be discussed in Section~\ref{sec:methods}, some fraction of the signals will be influenced by more than one tissue type, and therefore represent a mixture of classes.

A final clustering issue arises when computing centroids. In K-Means, cluster centroids are calculated as the mean of the points associated to each cluster, and in K-Medoids, the most centrally located data point is chosen to be the centroid. In both cases, high amounts of background noise, if included in any cluster, would strongly influence the mean, steering it away from a truer representation of the non-noise portion of the cluster.

Addressing these constraints has prompted the development of the method proposed in this paper, capable of learning features that relate characteristic signal shapes to individual components of the target, in a scalable and noise-robust manner. 

\section{Methods}
\label{sec:methods}

We wish to cluster TD signals based on their signal \emph{shape}, but not amplitude.
Generalizing from \Cref{eq:model}, a given pixel (and its corresponding TD signal) may be expressed in terms of characteristic signal shapes of one or more targets $\{ f_i \}$ and a residual term, $r_j(t)$,
\begin{equation}
    \TDInd{j} = \sum_{\forall i}\alpha_{i,j} f_i(t) + r_j(t) ,
        \label{eq:modelb}
\end{equation}
but where \emph{most} pixels come from only a single target, as had been the case in \Cref{eq:model}, such that $\{ \alpha_{i,j} \}$ is sparse. The proposed method is based on K-Means, but varies in its definition of distance and its method for computing cluster centroids (the learned features). Our goal is to discover characteristic signals shapes, a set of $K$ centroids, $\mathscr{F}=\{f_i(t)\}$, $i=1,...,K$, by creatively clustering the TD signals from a given image. 

Note that TD signals are just vectors in space, $\mathbb{R}^n$, where the dimension, $n$, of the space is simply the number of discrete TD samples. Thus, the equivalences $\TD \equiv \myVec{\TDAmp}, f(t) \equiv \myVec{f}$, \mbox{etc. are} made. Because TD signals are treated as Cartesian vectors, the signal shape is then analogous to the vector \emph{angle}. 

TD signals associated with a given target may be arbitrarily scaled and be subject to noise, thus any proposed clustering algorithm must be tolerant to these effects. The assumptions for the proposed method are as follows:
\begin{enumerate}
    \item Each underlying class (\ie isolated component) is characterized by a single prototype signal.
    \item \emph{Most} pixels are a member of only one class.  Relatively few pixels may be members of several classes (\ie exhibit a response associated with a mixture of underlying components).
    \item The noise level, both in background and in higher-amplitude signals, is significant.
    \item A large fraction of the pixels (the background) may be a member of {\em no} class, thus containing only noise, and thus should {\em not} be included in any centroid calculation.
\end{enumerate}

\subsection{Distance Metric}

The clustering algorithm requires a distance metric that is:
\begin{enumerate}
    \item Scale (amplitude) invariant, and
    \item Polarity-agnostic (\ie insensitive to signal inversions).
\end{enumerate}
Consider an arbitrary TD signal, $\myVec{\TDAmp}~=~m\myVec{u}$, for unit-vector $\myVec{u}$ that defines the characteristic signal. The \emph{negative} of this signal, $-\myVec{\TDAmp} = (-m)\myVec{u}$, shares the same direction, $\myVec{u}$, however the sign is opposite.  The angle between $\myVec{\TDAmp}$ and $-\myVec{\TDAmp}$ is $\pi$ radians, however they share the exact same signal shape, and are thus associated with the same underlying imaged target. Therefore, a polarity-agnostic distance metric must assign a distance of zero between  $\myVec{\TDAmp}$ and $-\myVec{\TDAmp}$.

For simplicity, and to achieve symmetry in the distance metric, all angles will be considered to be positive. The angle, $\vartheta$, between two vectors, $\myVec{v}_1$ and $\myVec{v}_2$, is defined as
\begin{equation}
    \vartheta := \angle\left(\myVec{v}_1, \myVec{v}_2\right) = \arccos \left( \frac{\langle\myVec{v}_1, \myVec{v}_2\rangle}{\lVert\myVec{v}_1\rVert \lVert\myVec{v}_2\rVert} \right),
\end{equation}
for $\arccos(\cdot)$ defined abstractly as $x \mapsto \arccos(x)$, $[-1,1]~\to~[0,\pi]$.
The proposed distance metric, satisfying polarity-agnosticism, is 
$d\left(\myVec{v}_1, \myVec{v}_2\right) = \sin(\vartheta),$
for $\sin(\cdot)$ defined on the interval $[0,\pi]$, constraining the range of $d(\cdot)$ to $[0,1]$. Note that the function composition $\sin\left(\arccos(x)\right) = \sqrt{1 - x^2},$
therefore the distance metric simplifies as
\begin{equation}
    d(\myVec{v}_1, \myVec{v}_2) = \sqrt{1 - \left( \frac{\langle\myVec{v}_1, \myVec{v}_2\rangle}{\lVert\myVec{v}_1\rVert \lVert\myVec{v}_2\rVert} \right)^2}.
\end{equation}

\subsection{Calculation of Cluster Centroids}

As was mentioned at the end of \Cref{sec:background}, an alternative method for computing cluster centroids is required. Although angle-based variations of K-Means do exist --- such as Spherical K-Means \cite{dhillon2001concept,nguyen2008gene,hornik2012spherical} whereby data-points are projected onto the unit-hypersphere via normalization and distance is defined by cosine \emph{dissimilarity} --- centroids are still computed by taking the mean of all points within a given cluster, an approach which does not apply to \Cref{eq:modelb}:
\begin{itemize}
    \item We have a requirement for polarity-agnosticism, whereby antipodal TD signals are clustered together, and these signals will largely cancel (negate each other) if averaged.
    \item A large fraction of pixels are background (composed of zero-mean noise), having a random angle, and therefore are associated to clusters at random, leading to significant biases if included in the sample mean, and it is highly undesirable for background noise to dominate (or affect at all) the learned centroids.
\end{itemize}
Similarly, while angular distance metrics can be used with K-Medoids, centroids are selected to minimize intra-cluster distance and would be \emph{strongly} influenced by large fractions of background noise (random angles).
Instead, we desire a unit-vector pointing in the direction of the non-noise portion of the given cluster to define the centroid. 
The method needs to be polarity-agnostic, however we do not actually know, a priori, whether both positive and negative examples are present. Given the set of points, $\mathscr{S}_i$, associated with cluster, $i$, we construct
the union set,
\begin{equation}
\mathscr{S}_i^{\pm} = \mathscr{S}_i \bigcup\,\left(-\mathscr{S}_i\right),
\end{equation}
made up of the cluster, $\mathscr{S}_i$, and its negated points, $-\mathscr{S}_i$.
From $\mathscr{S}_i^{\pm}$, the centroid can be found as the direction of greatest variance (the first principal component from the sample covariance of $\mathscr{S}_i^{\pm}$), allowing higher amplitude signals (having a greater signal to noise ratio) to have greater influence, effectively eliminating the influence of background noise (those random-angle data-points near the origin).


\subsection{Algorithm}

With the distance metric and the method for calculating centroids now defined, the clustering algorithm, detailed in \Cref{alg:NickKMeans}, follows fairly naturally from conventional \mbox{K-Means}. The calculation of cluster centroids is detailed in Centroid Update (\cref{line:NickKMeans_CentroidUpdateBlock}), and on \cref{line:NickKMeans_SVD_step} an SVD (Singular Value Decomposition) is used to extract the first principal component.

Similar to conventional K-Means, we need to define convergence criteria:
\begin{enumerate}
    \item Sufficiently few data-points change clusters between iterations (algorithm parameter  $\textit{MovesCriterion}$), or
    \item The difference in the mean residual is sufficiently small between iterations (algorithm parameter $\textit{DifferenceCriterion}$).
\end{enumerate}

Feature amplitude extraction is achieved by performing a change of basis on the TD signals. Given the set of feature vectors, $\mathscr{F}=\{\myVec{f}_i\}$, resulting from \Cref{alg:NickKMeans}, the TD signals can be expressed as a weighted sum of the feature vectors plus a residual term:
\begin{equation}
    \TDInd{j} = \sum_{\forall i}{\alpha_{i,j} \myVec{f}_i} + \myVec{r}_j.
\end{equation}
Arranging the feature vectors as matrix columns, forming a matrix of features, $F = \big[\myVec{f}_1 \mid \myVec{f}_2 \mid ...\big]$,
the preceding expression then becomes $\TDInd{j} = F \myVec{\alpha}_j + \myVec{r}_j$.
To solve for the vector of feature weights, $\myVec{\alpha}$, the pseudo-inverse \cite{fieguth2010statistical} of $F$ is used, thus, $\myVec{\alpha}_j = F^{+} \TDInd{j}$.

\begin{algorithm}[!t]
    \caption{Proposed Clustering Algorithm}\label{alg:NickKMeans}
    
    \Input{Set of TD signals, $\mathscr{S}=\{\TDInd{j}\}$, to be clustered.\newline
    Number of desired clusters, $K$.\newline
    Minimum allowable moves criterion, $\textit{MovesCriterion}$. \newline
    Difference in mean residual criterion, $\textit{DifferenceCriterion}$.}
    \Output{Set of cluster labels, $\mathscr{L}=\{\ell\}$, associated with each TD signal.\newline
    Set of cluster centroids, $\mathscr{F}=\{\myVec{f}_i\}$, for $i=1,...,K$.}
    
    \begin{algorithmic}[1] 
        \Comment{\underline{Initialization}:}
        \Comment{Randomly select $K$ data-points as initial centroids.}
        \For{$i = 1,...,K$}
            \State $\myVec{f}_i \overset{Random}{\gets} \TD \in \mathscr{S}$
        \EndFor
        \Comment{Set previous value of mean residual to 0.}
        \State $\mu_r^{prev} \gets 0$
        
        \Repeat
            \FirstLineComment{Set number of changed cluster labels to 0.}
            \State $moves \gets 0$
            \vspace{0.25cm}
            \Comment{\underline{Membership Update}: Finding nearest centroid to each point.}
            \ForAll{$\TDInd{j} \in \mathscr{S}$}
                \State $\ell_j \gets \argmin_{\,i\in\{1,...,K\}} \Big\{d\Big(\TDInd{j},\myVec{f}_i\Big)\Big\}$
                \Comment{Increment $moves$ if cluster membership changes.}
                \If{$\ell_j$ changed this iteration}
                    \State $moves \gets moves + 1$
                \EndIf
            \EndFor
            \vspace{0.25cm}
            \Comment{Evaluate mean residual (objective).}
            \State $\mu_r \gets \frac{1}{\lVert\mathscr{S}\rVert} \sum_{\TDInd{j} \in \mathscr{S}}{\, d\left(\TDInd{j},\myVec{f}_{\ell_j}\right)} $
            \vspace{0.1cm} 
            \State $\Delta\mu_r \gets \mu_r - \mu_r^{prev}$
            \State $\mu_r^{prev} \gets \mu_r$
            \vspace{0.25cm}
            \Comment{\underline{Centroid Update}: Use data-points within clusters to update.}
            \For{$i = 1,...,K$}
                \label{line:NickKMeans_CentroidUpdateBlock}
                \FirstLineComment{Get set of data-points within cluster.}
                \State $\mathscr{S}_{i} \gets \left\{\TDInd{j} \,\middle|\, \ell_j = i\right\}$
                \Comment{Take union of set with its negative.}
                \State $\mathscr{S}_i^{\pm} \gets \mathscr{S}_i \bigcup\,\left(-\mathscr{S}_i\right)$
                \Comment{Compute first principal component via SVD. Assign to centroid.}
                \State $\myVec{f}_i \gets \text{PC}_1\left(\mathscr{S}_i^{\pm}\right)$
                \label{line:NickKMeans_SVD_step}
                \Comment{Normalize centroid to fall on unit-hypersphere}
                \State $\myVec{f}_i \gets \myVec{f}_i / \lVert \myVec{f}_i \rVert$
            \EndFor
        \Until{$\Delta\mu_r \le \textit{DifferenceCriterion}$ \textbf{OR} \newline $moves \le \textit{MovesCriterion}$}
    \end{algorithmic}
\end{algorithm}

\section{Results}

\subsection{Comparison of Methods on Synthetic Data}
\label{subsec:synth_results}


We begin by comparing the proposed method to standard Principal Components Analysis (PCA) and  Angular-Distance K-Means on the basis of synthetic data.

On the basis that none of the methods being compared are sensitive to correlations between specific indices / samples of the TD signals, randomly generated synthetic data are used. Additionally, it is the \emph{angle} between vectors / signals that matters.  It can be verified that the \emph{ordering} of indices has no impact on angle by considering permutation matrices \cite{zhang2011matrix}, which can be used to permute the signal / vector indices, but do not cause the \emph{angle} between vectors to change because permutation matrices are orthogonal. Ground truth prototype signals are first generated, then scaled (including negatives) versions with added noise are used to generate synthetic data. Additionally, background noise (zero-signal with added noise) is included.  To improve visual interpretability, to appear similar to TD signals, temporal correlations are introduced to the prototypes using a first-order auto-regressive model \cite{kirchgassner2012introduction}.

\begin{figure}[t]
    \centering
    \includegraphics[width=0.9\columnwidth,trim={0cm 0.25cm 0cm 0.425cm},clip]{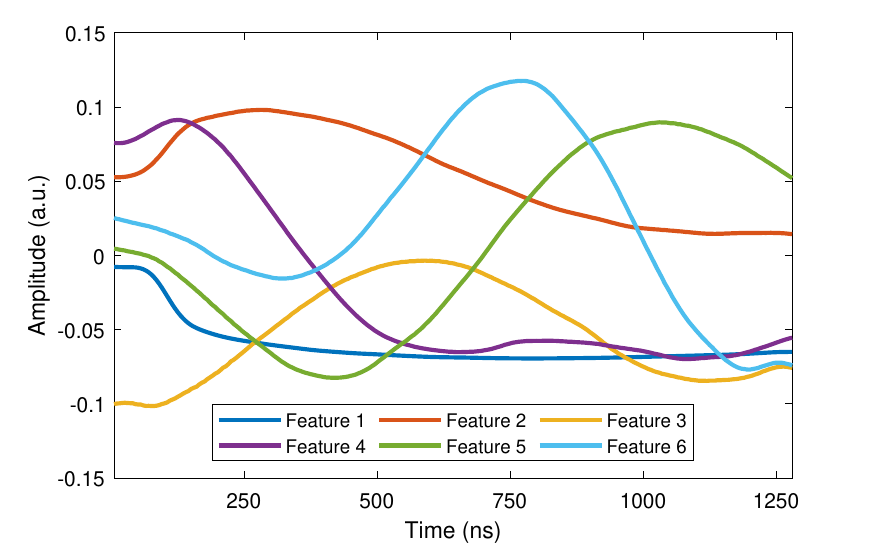}
    \caption{The cluster centroids learned from the human breast tissue slide in \Cref{fig:K_means_original_breast_tissue}. A wide variety of signal shapes are learned and presented here.}
    \label{fig:K_means_breast_tissue_centroids}
\end{figure}

\begin{figure*}[tp]
\centering
\resizebox{0.95\textwidth}{!}{
\scriptsize
\begin{tabular}{ccccc}
    \ctab{} & 
    \hspace{-0.5cm}\ctab{Ground\\ Truth} & 
    \hspace{-0.5cm}\ctab{PCA} & 
    \hspace{-0.5cm}\ctab{Angular-Distance\\ K-Means} &
    \hspace{-0.5cm}\ctab{Proposed\\ Method} \\

    \ctab{\rotatebox{90}{\parbox{3cm}{\centering Prototypes~/\\Learned Features}}} & 
    \hspace{-0.75cm}\ctab{\includegraphics[width=0.23\textwidth,trim={0 0 1.1cm 0.5cm},clip]{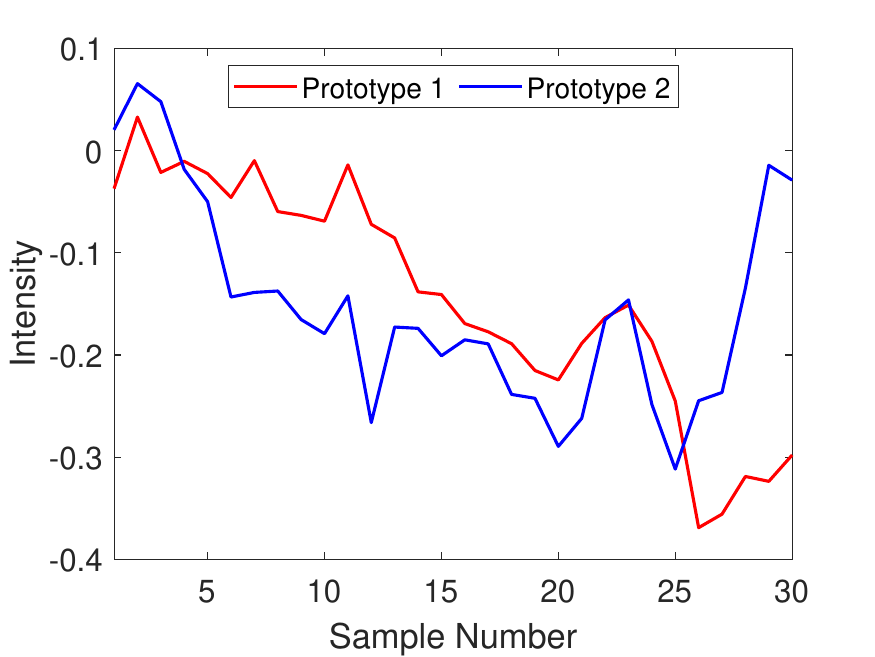}} & 
    \hspace{-0.75cm}\ctab{\includegraphics[width=0.23\textwidth,trim={0 0 1.1cm 0.5cm},clip]{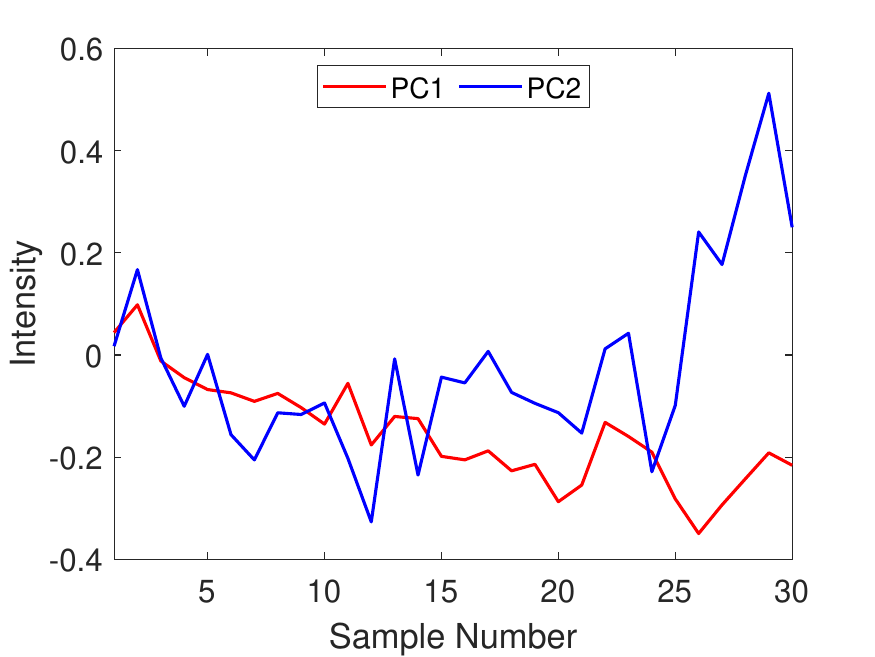}} &
    \hspace{-0.75cm}\ctab{\includegraphics[width=0.23\textwidth,trim={0 0 1.1cm 0.5cm},clip]{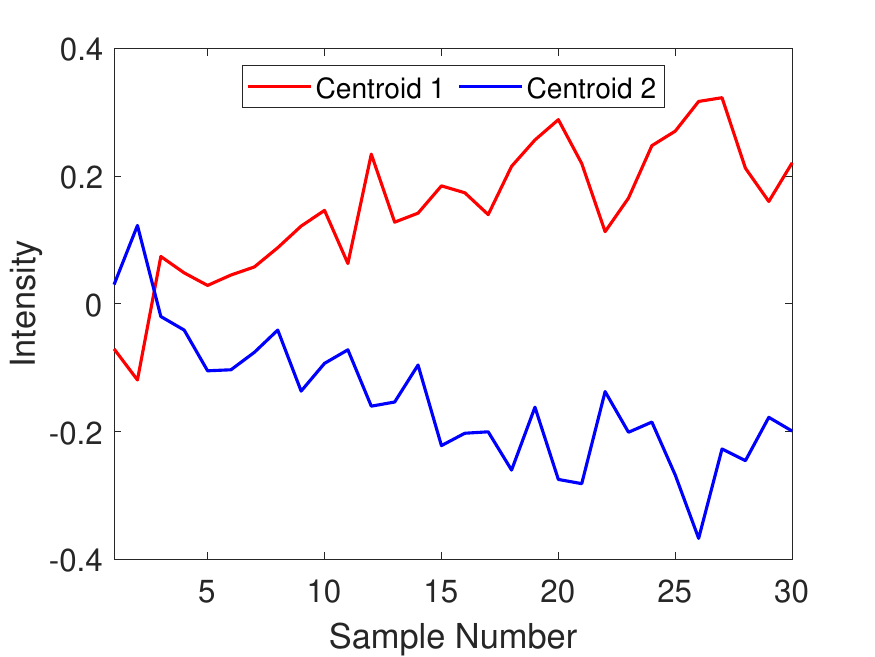}} & 
    \hspace{-0.75cm}\ctab{\includegraphics[width=0.23\textwidth,trim={0 0 1.1cm 0.5cm},clip]{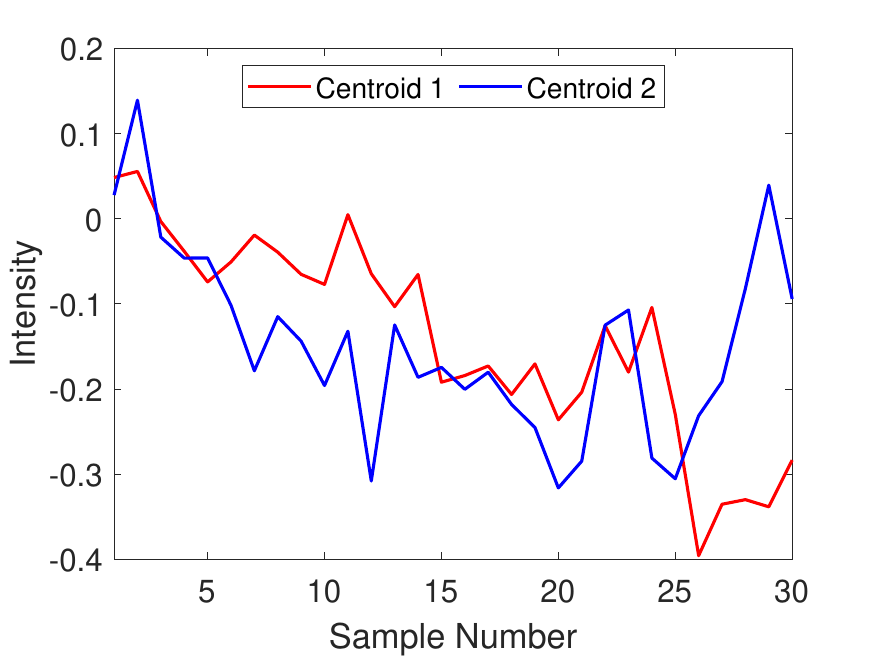}} 
    \vspace{-0.08cm}\\
    
    \vspace{-0.2cm}
    \ctab{\rotatebox{90}{\parbox{3cm}{\centering Data-Points~\&\\Membership}}} & 
    \hspace{-0.75cm}\ctab{\includegraphics[width=0.23\textwidth,trim={0 0 1.1cm 0.5cm},clip]{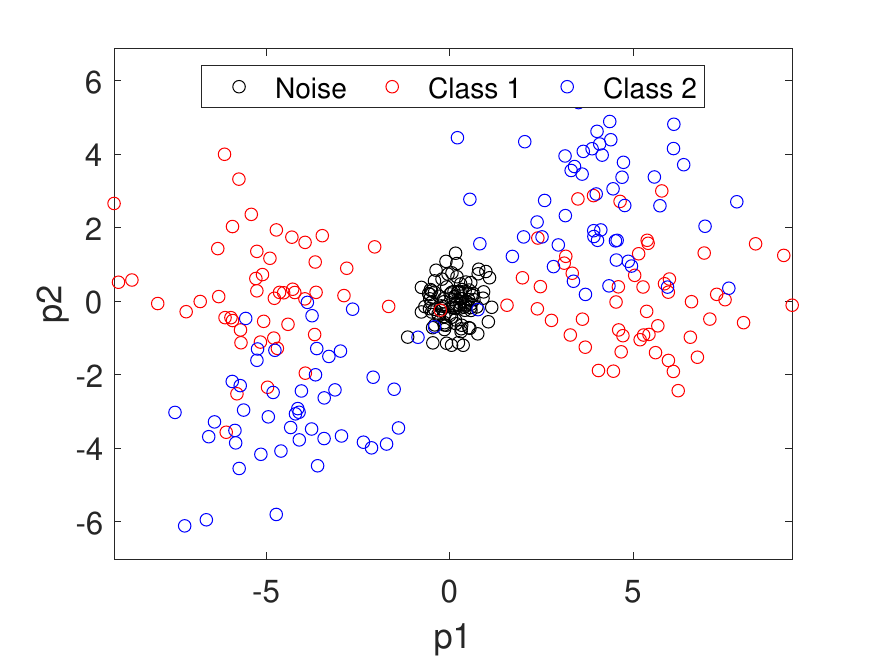}} &
    \hspace{-0.75cm}\ctab{\includegraphics[width=0.23\textwidth,trim={0 0 1.1cm 0.5cm},clip]{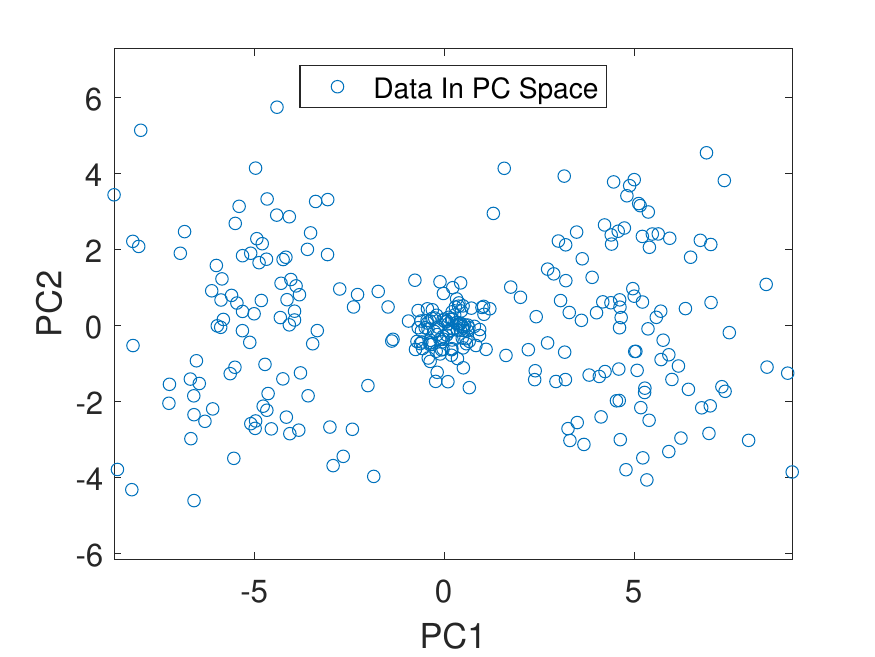}} &
    \hspace{-0.75cm}\ctab{\includegraphics[width=0.23\textwidth,trim={0 0 1.1cm 0.5cm},clip]{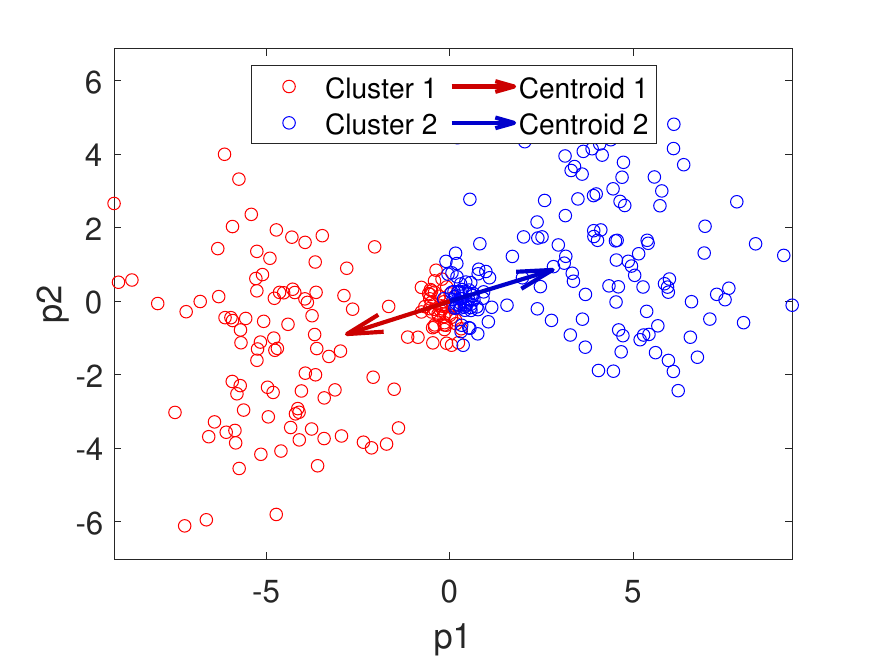}} &
    \hspace{-0.75cm}\ctab{\includegraphics[width=0.23\textwidth,trim={0 0 1.1cm 0.5cm},clip]{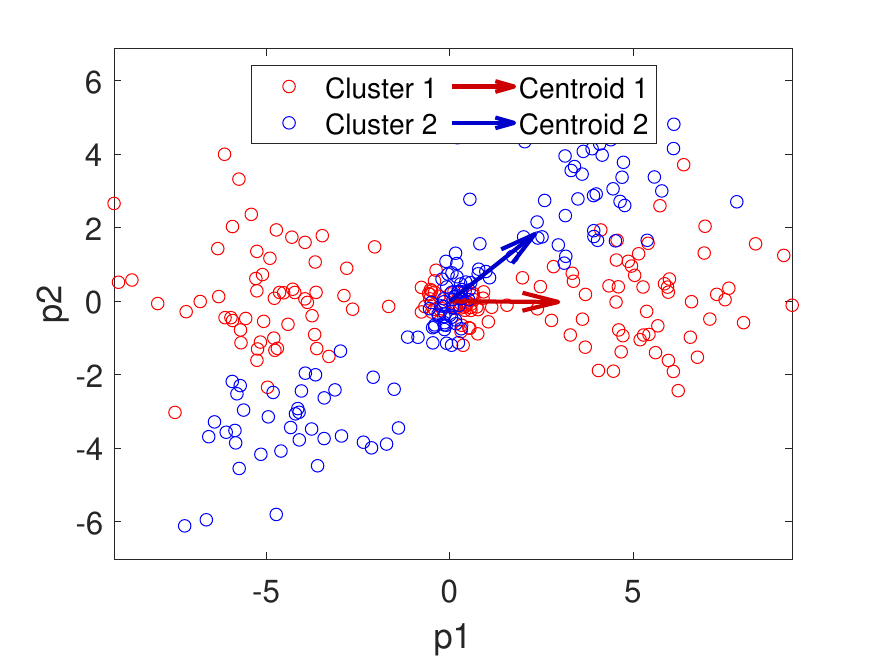}}
\end{tabular}
}
\caption{Various methods are applied to learn features from synthetic data. The first column shows the ground truth classes present in the data and the associated true cluster prototypes. Note that the data-points, shown in the bottom row, are actually 30-dimensional but are projected onto a plane defined by true prototypes, with planar coordinates denoted p1 and p2, for ease of viewing (except for the PCA case, where a projection into the principal component space is used instead). In addition to the two well-defined classes, there is low amplitude background noise present in the data set. The following columns demonstrate the ability of Angular-Distance K-Means, PCA, and the proposed method to recover the true clusters and centroids. In the case of PCA, no such clusters are identified; however, the principal components (learned features) can be compared to the true prototypes.}
\label{fig:synthetic_results}
\end{figure*}

The synthetic data tests are illustrated in \Cref{fig:synthetic_results}. The first column contains the ground truth prototypes (top) and class membership of the data-points (bottom). For visualization purposes in the bottom row, the data-points, which are 30-dimensional, are projected onto a 2D plane defined by the ground truth prototypes (except for the case where PCA is used, where the data-points are projected into the principal component space). 

It can clearly be seen that neither PCA nor the Angular-Distance K-Means produce learned features that even resemble the ground truth prototypes, since neither approach is able to produce the desired sparse-weighted features, whereas the proposed method is in fact highly effective at doing so. 

\subsection{Quantitative Evaluation of Proposed Method}
\label{subsec:quant_eval}

To more comprehensively test the proposed approach, the synthetic test of \Cref{subsec:synth_results} is generalized by evaluating performance as a function of key problem parameters, including the nominal angular separation of the underlying classes, the noise level, the fraction of points of low amplitude (background noise), and the fraction of points that are a mixture of multiple classes.
Synthetic data (sets of 500 points) comprising two classes are generated in 1024-dimensions, mimicking real PARS TD data, based on ground truth class prototypes with a specified angular separation. To form a comparison, the proposed method,  Spherical K-Means, Angular-Distance K-Means, and PCA are evaluated using this input data.  Problem parameters are varied individually, holding all others fixed at a given nominal value, and results are averaged over 50 trials to mitigate error.

\begin{figure*}[tp]
\centering
\resizebox{0.9\textwidth}{!}{
\scriptsize
\begin{tabular}{cccc}
    \hspace{0.2cm}\ctab{Angular Separation} & 
    \hspace{-0.3cm}\ctab{Noise Level} & 
    \hspace{-0.3cm}\ctab{Background Noise\\Fraction} &
    \hspace{-0.3cm}\ctab{Mixed-Class\\Fraction} \\[-0.75ex]

    \hspace{-0.25cm}\ctab{\includegraphics[width=0.23\textwidth,trim={0 0 1.1cm 0.5cm},clip]{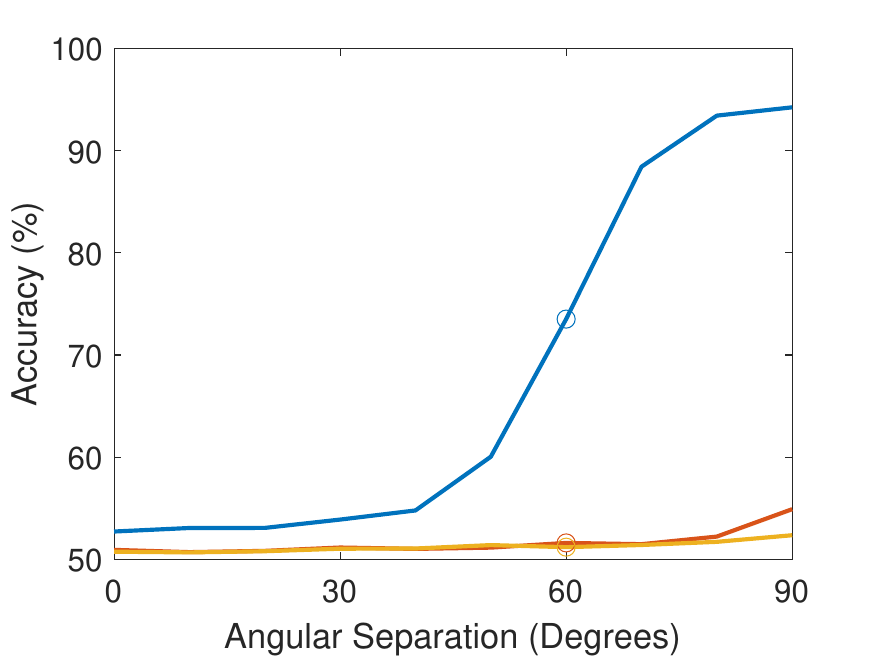}} & 
    \hspace{-0.75cm}\ctab{\includegraphics[width=0.23\textwidth,trim={0 0 1.1cm 0.5cm},clip]{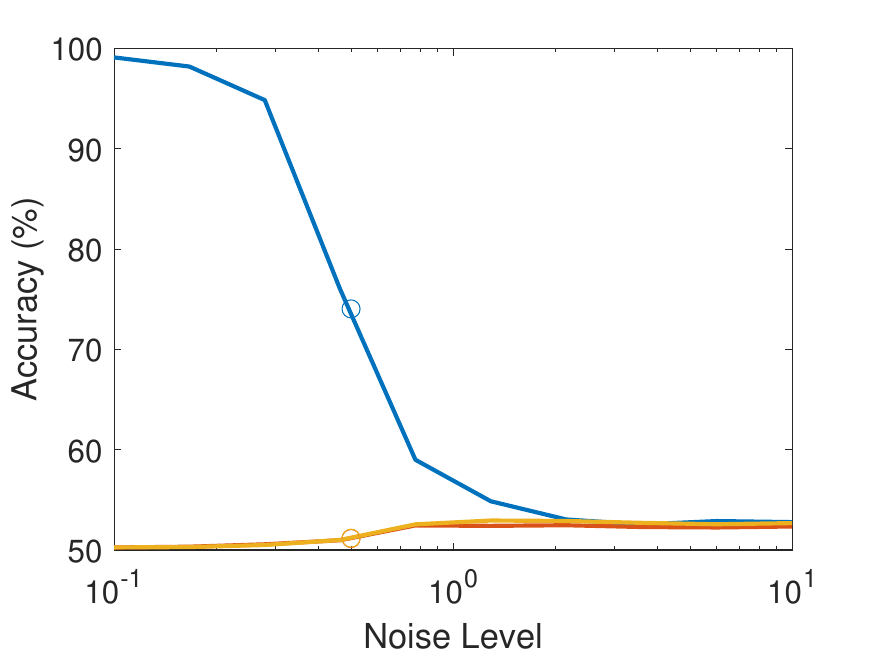}} &
    \hspace{-0.75cm}\ctab{\includegraphics[width=0.23\textwidth,trim={0 0 1.1cm 0.5cm},clip]{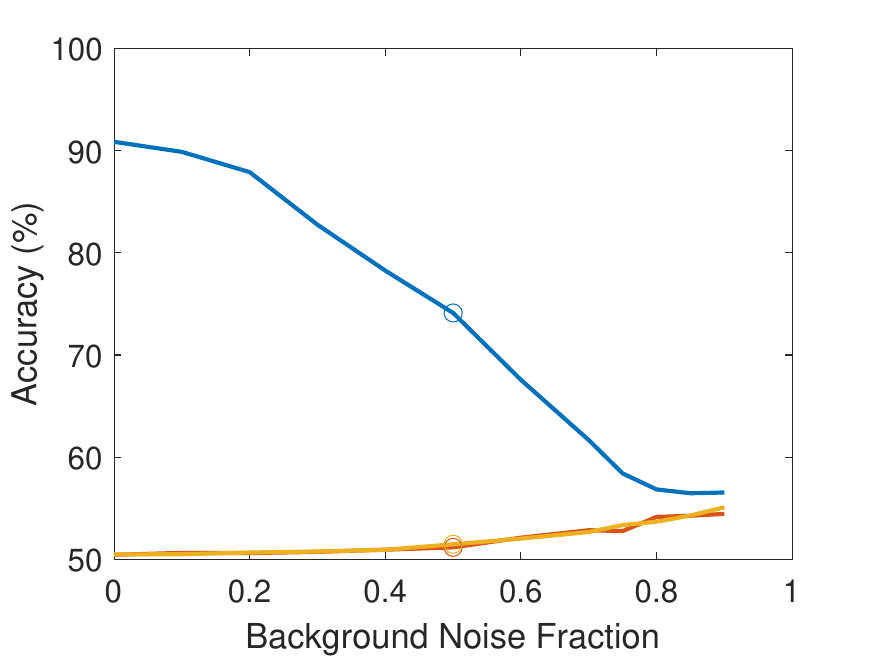}} & 
    \hspace{-0.75cm}\ctab{\includegraphics[width=0.23\textwidth,trim={0 0 1.1cm 0.5cm},clip]{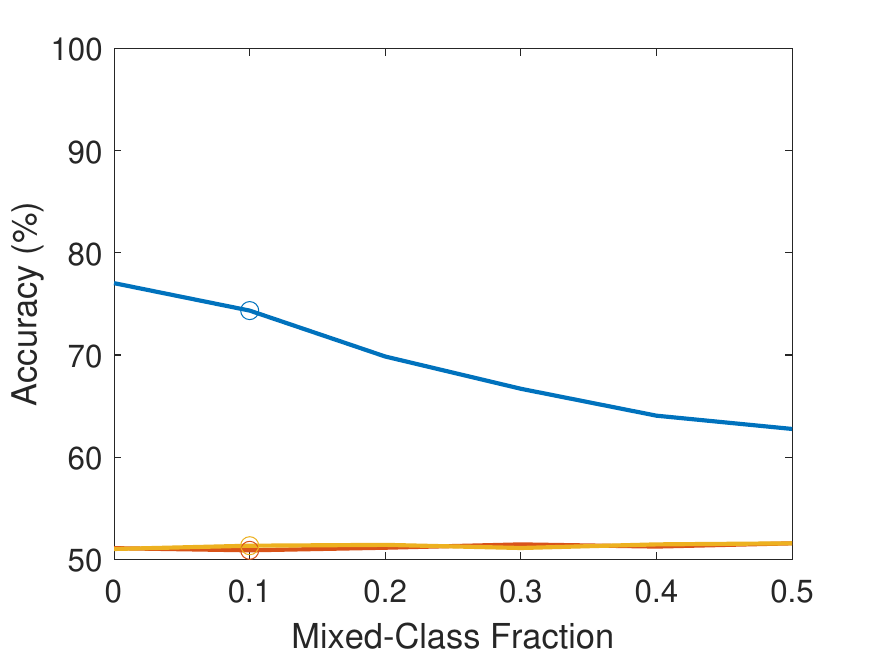}} 
    \\[-0.5ex]
    
    \hspace{-0.25cm}\ctab{\includegraphics[width=0.23\textwidth,trim={0 0 1.1cm 0.5cm},clip]{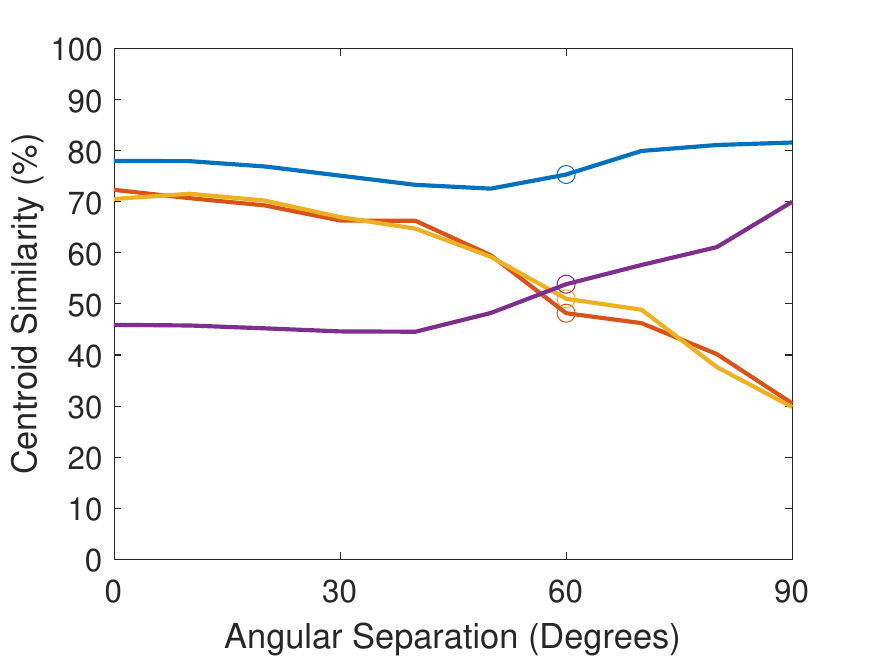}} &
    \hspace{-0.75cm}\ctab{\includegraphics[width=0.23\textwidth,trim={0 0 1.1cm 0.5cm},clip]{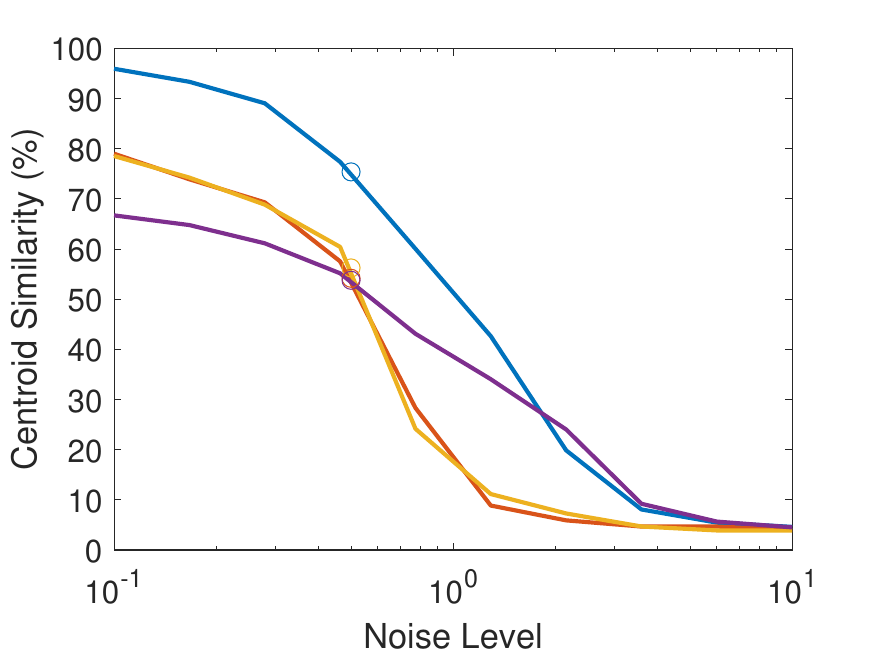}} &
    \hspace{-0.75cm}\ctab{\includegraphics[width=0.23\textwidth,trim={0 0 1.1cm 0.5cm},clip]{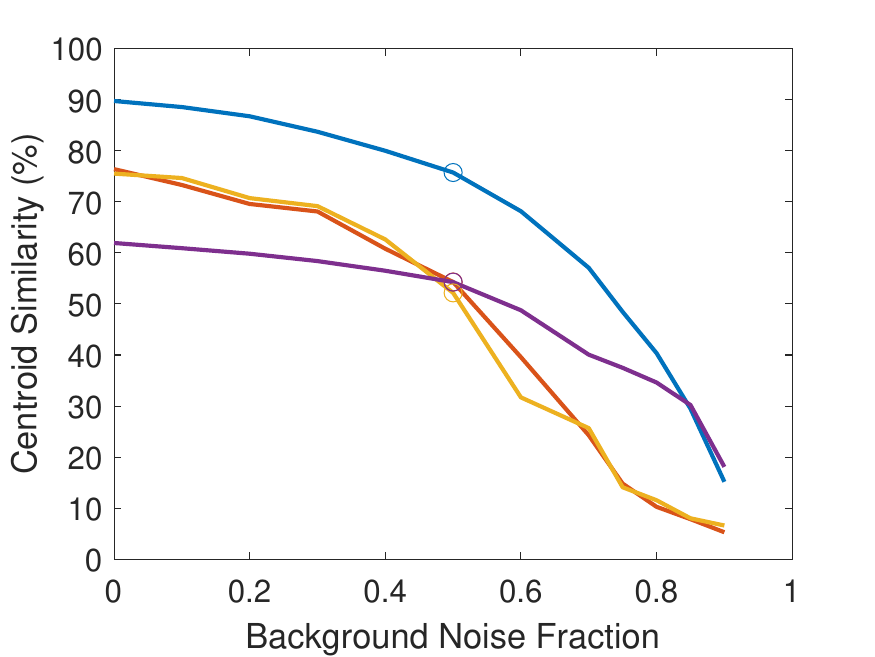}} &
    \hspace{-0.75cm}\ctab{\includegraphics[width=0.23\textwidth,trim={0 0 1.1cm 0.5cm},clip]{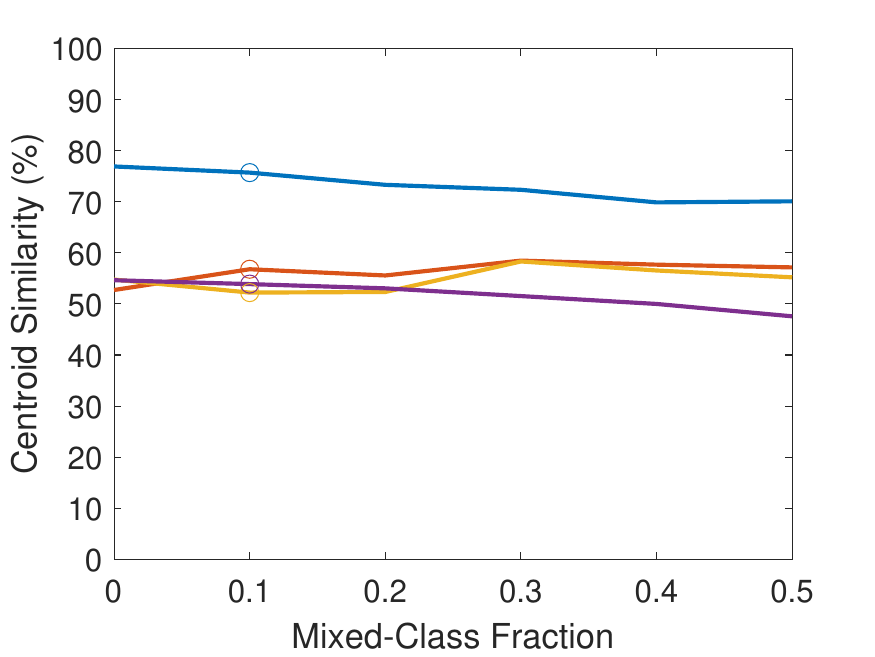}}
    \\
    
    \multicolumn{4}{c}{\hspace{0.5cm}\ctab{\includegraphics[width=0.6\textwidth]{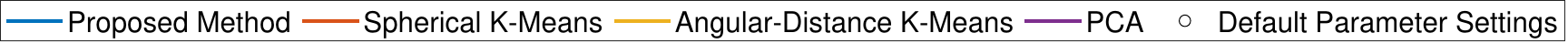}}}
\end{tabular}
}
\caption{
For synthetically generated data in 1024-dimensions, the proposed method, Spherical K-Means, Angular-Distance K-Means, and PCA are used to cluster the data and learn features~/ centroids. In all cases, two features are learned. With PCA, points are not clustered; however, principal components are learned. 
By comparing to the ground truth classes and prototypes, the clustering accuracy (top row) and centroid similarity (bottom row) are evaluated for each method. 
In each column, the performance of the methods is evaluated while varying one parameter of the data.
Results are averaged over 50 trials.
In each column only one parameter is varied, holding the other parameters at their default values. Results at the default parameter values are circled.
}
\label{fig:method_comparison_parameter_evaluation}
\end{figure*}

Results are shown in \Cref{fig:method_comparison_parameter_evaluation}. 
In all cases, two ($K=2$) features are learned. With PCA, points are not clustered; however, principal components are learned. 
By comparing to the ground truth classes and prototypes, the clustering accuracy (top row) and centroid similarity (bottom row) are evaluated for each method. 
In each column, the performance of the methods is evaluated independently for each parameter.
It can clearly be observed that the proposed method significantly outperforms the other three tested methods in nearly all cases. 
Further discussion is given in \Cref{sec:discussion}.

\subsection{Application Study --- PARS Data}
\label{subsec:real_PARS_results}

Finally, the proposed method is applied to real-world data collected from a PARS microscope. Centroids (representative features) are learned and then are used to extract feature amplitudes from the TD signals of the PARS image. 

\begin{figure*}[!t]
    \centering
    \includegraphics[width=0.9\textwidth]{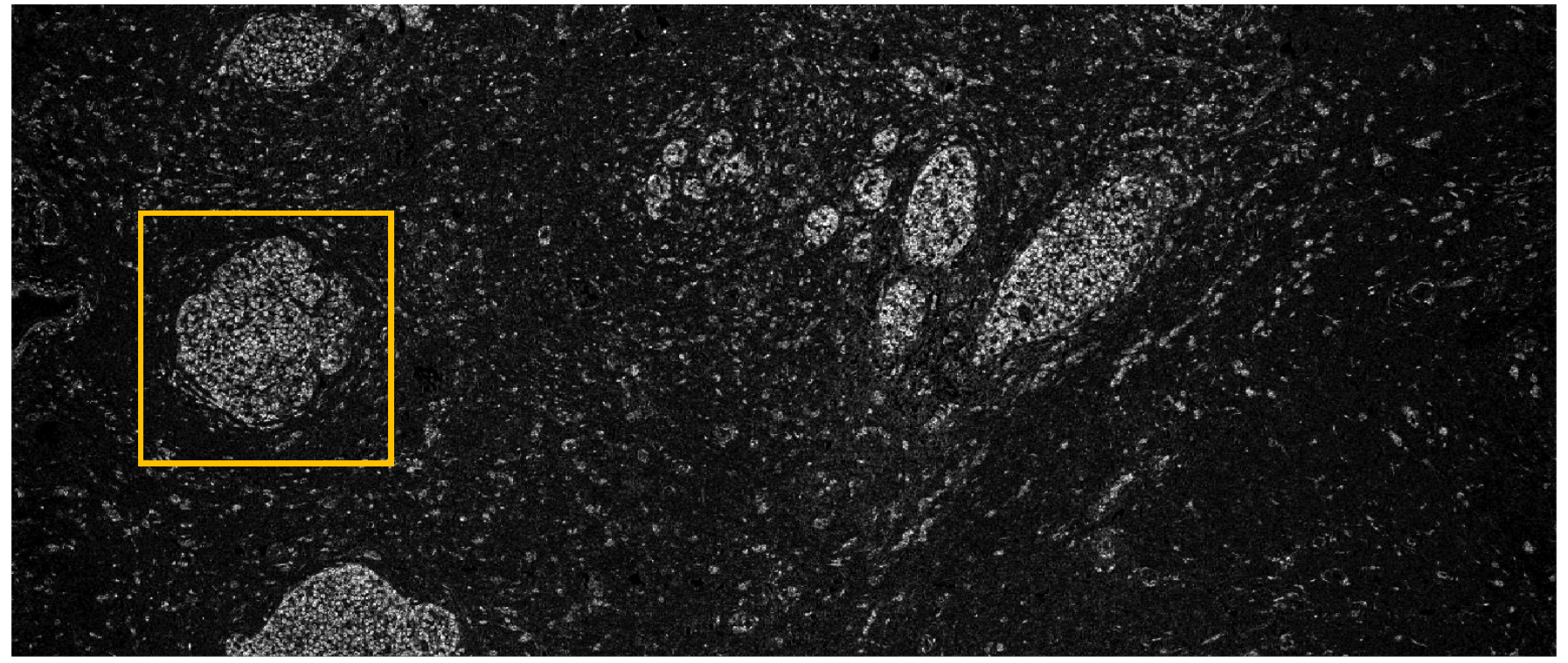}
    \caption{A standard scalar signal projection of PARS imagery of unstained human breast tissue on a slide.
    The time-domain signals from the yellow boxed region are used for feature learning. Image captured by Benjamin R.~Ecclestone.}
    \label{fig:K_means_original_breast_tissue}
\end{figure*}

The use of the clustering algorithm for performing feature extraction is demonstrated here on an unstained, formalin-fixed paraffin-embedded (FFPE) human breast tissue slide. The standard projection PARS image of the slide is shown in \Cref{fig:K_means_original_breast_tissue}. This image was captured and provided by Benjamin R.~Ecclestone with gratitude from the authors. 
A boxed-in region indicates the selection of TD signals that were used for feature learning, and an input of $K=6$ desired clusters was used. The task of selecting the number of clusters, $K$, is not addressed in this paper, and in principle the same issues apply as in selecting $K$ in K-Means or dimensionality in PCA, and is assumed to be known based on external information.

\begin{figure*}[!t]
    \centering
    \includegraphics[width=0.9\textwidth]{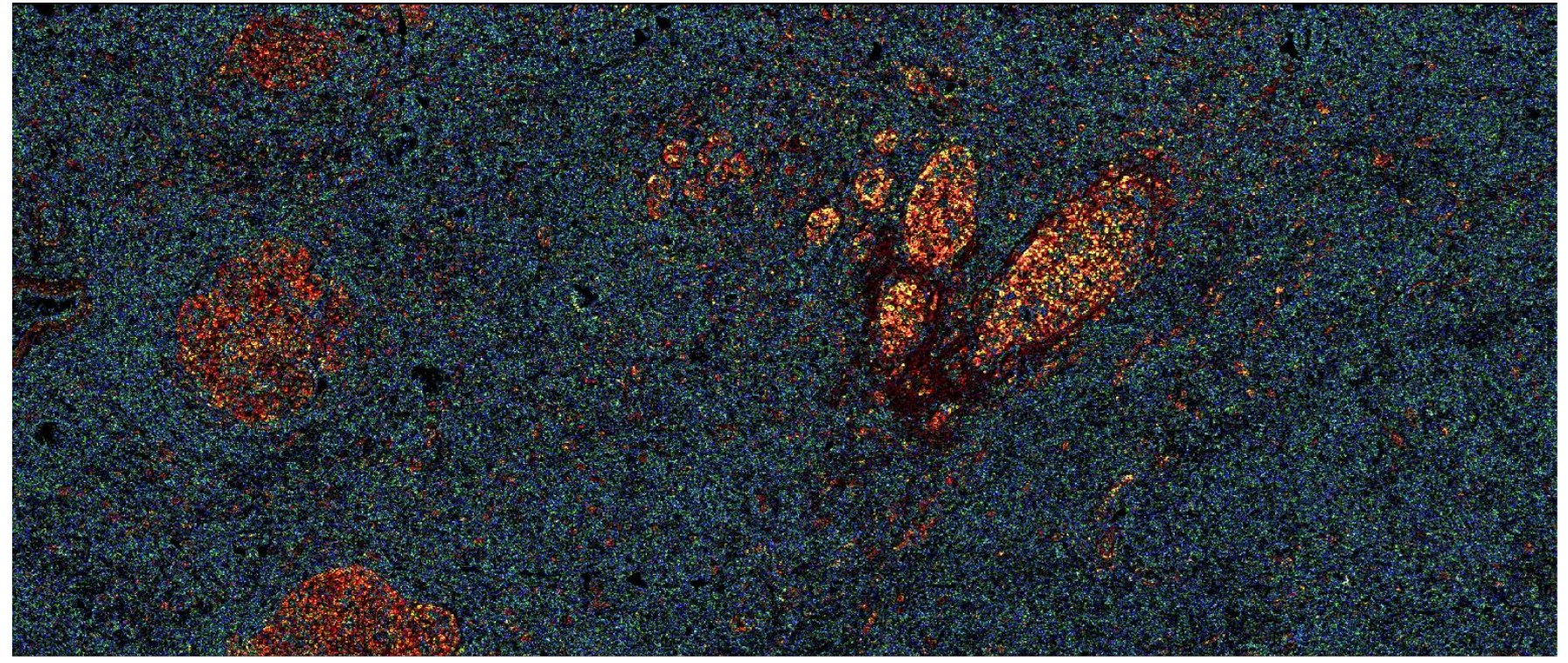}
    \caption{Absolute values of three feature amplitudes, extracted from the time domain signals underlying the imagery in \Cref{fig:K_means_original_breast_tissue}.  The three features are combined to produce a colour image:  Feature 1 maps to \textcolor{red}{red}, Feature 4 maps to \textcolor{green}{green}, Feature 5 maps to \textcolor{blue}{blue}. Quite remarkable structures and tissue differentiation are clearly visible in this image, relative to that of \Cref{fig:K_means_original_breast_tissue}.}
    \label{fig:K_means_breast_tissue_feature_combo_RGB}
\end{figure*}

The learned centroids are shown in \Cref{fig:K_means_breast_tissue_centroids}. A wide variety of signal shapes are present, each centroid representing a unique tissue type.
The extracted feature amplitudes (in absolute value, thus ignoring the effects of polarity) are combined as an RGB image in \Cref{fig:K_means_breast_tissue_feature_combo_RGB}, with the amplitude of three features mapped into RGB space, thus Feature 1 maps to red, feature 4 maps to green, and feature 5 maps to blue, further discussed in \Cref{sec:discussion}.

\section{Discussion}
\label{sec:discussion}

The comparison between methods, shown in \Cref{fig:synthetic_results}, illustrates qualitatively that the learned features of PCA (the principal components) and Angular-Distance K-Means do not adequately match the ground truth. This can be explained by considering the underlying objective of each method:
\begin{itemize}
    \item PCA seeks to learn a minimal basis (set of principal components) that preserves maximal variation in the data, but makes no assumption about the underlying basis nor the associated sparsity of its weights. While the principal components could, in principal, be a linear combination of the true prototypes, information of the true prototypes cannot directly be inferred. 
    \item Regular K-Means, based on a Euclidean distance in the native dimensional space, fails to assert the appropriate distance metric, and so is unable to accommodate variable signal amplitudes and would, in any event, most likely create a separate class for low-amplitude noise signals.
    \item Angular-Distance K-Means creates clusters such that minimal intra-cluster variance and maximal inter-cluster distance are achieved, as is intended; however, due to the bi-polar nature of the TD signals explored in this report, Angular-Distance K-Means cannot group antipodal components of the same class together.  Even if the distance metric were changed to accommodate such bi-polar samples,  \mbox{K-Means} still remains highly susceptible to the influence of background noise since its centroids are not weighted by signal amplitude. 
\end{itemize}
Regarding the quantitative evaluations shown in \Cref{fig:method_comparison_parameter_evaluation}, in general, most of the trends match intuition --- decreasing performance is seen with higher noise levels or more mixed-class points and increasing performance with greater separation angles. 
Notably, the proposed method is able to learn centroids well at low separation angles in spite of its poor clustering accuracy, demonstrating its ability to learn and discern exceedingly similar underlying features.
Spherical and Angular-Distance K-Means perform extremely similarly.
Somewhat counter-intuitively, these two methods work well with \emph{lower} angular separation and \emph{greater} fractions of background and mixed points. This is because the methods tend to learn two \emph{antipodal} centroids, half way between the two ground truth centroids, as was seen in \Cref{fig:synthetic_results}. By having low angular separation and high fractions of background and mixed-class points, the methods can more easily learn centroids that fall directly between the ground truth ones.

Although ground truth is not available for the PARS imagery, the applied use of the proposed method yielded biologically meaningful, spatially grouped, highly compelling results.  The learned features shown in \Cref{fig:K_means_breast_tissue_centroids} represent a wide variety of signal shapes which correspond to specific tissue structures present in the image, and distinctly separated in a way that is not the case in standard PARS imagery.  A further  detailed study involving commentary from histology experts would be necessary to validate the inferred features, and is the subject of ongoing research.

\section{Conclusion}

The method proposed in this paper is capable of learning features from high-dimensional signals that relate to individual components of the underlying data, in spite of signal amplitude variations, inversions, and noise. When tested on synthetic data, the proposed method learned features that closely match the ground truth prototypes, whereas the other compared methods could not.

The proposed method consistently performed as well or better than the other methods, across all four problem parameters, 
offering an attractive, intuitive, amplitude-flexible and infrequent-class-mixing alternative to PCA or \mbox{K-Means}, due to its definition of distance, mixing weight sparsity, and strategy for computing centroids.

When applied to real data from a PARS microscope, the proposed method yielded features that showed correspondence with tissue structures, showing significant promise.  Clearly the proposed method is not specific to PARS data, and is far more broadly applicable to \mbox{multi-dimensional / multi-modal} signals than just those explored here.


\section*{Acknowledgment}

The authors would like to thank Benjamin R.~Ecclestone, a PhD student at PhotoMedicine Labs at the University of Waterloo, for collecting the PARS image analyzed here.

\IEEEtriggeratref{25} 


\bibliographystyle{IEEEtran}
\bibliography{IEEEabrv,bib/mybib}
%



\end{document}